\documentclass[letterpaper]{article} 
\usepackage{aaai2027}  
\nocopyright 
\usepackage[hyphens]{url}  
\usepackage{graphicx} 
\usepackage{natbib}  
\usepackage{caption} 
\usepackage{amsmath}
\usepackage{amssymb}
\usepackage{booktabs}
\usepackage{multirow}
\usepackage{array}
\usepackage{colortbl}

\newcommand{\nohint}{\textsc{Blind}}
\newcommand{\hint}{\textsc{Leaked}}
\newcommand{\afr}{\mathrm{AFR}}

\title{Answer-Conditioned Chains of Thought Degrade Verifiable-Reasoning Distillation in Large Language Models}

\author{
    Jungseob Lee\textsuperscript{\rm 1},
    Seungyoon Lee\textsuperscript{\rm 1},
    Suhyune Son\textsuperscript{\rm 1},
    Dongyub Jude Lee\textsuperscript{\rm 2},
    Sungbin Han\textsuperscript{\rm 1},\\
    Sugyeong Eo\textsuperscript{\rm 3}\corresponding,
    Heuiseok Lim\textsuperscript{\rm 1}\corresponding
}
\affiliations{
    \textsuperscript{\rm 1}Korea University\\
    \textsuperscript{\rm 2}Zoom Communications\\
    \textsuperscript{\rm 3}Yonsei University\\
    \{omanma1928, dltmddbs100, ssh5131, sungbinhan9039, limhseok\}@korea.ac.kr, jude.lee@zoom.us, s.eo@yonsei.ac.kr
}

\begin{document}

\maketitle

\begin{abstract}
A standard recipe for distilling the reasoning ability of large language models (LLMs) is to sample chains of thought from the model, keep those that reach the correct final answer, and fine-tune on the survivors. When sampling fails, a common fix shows the generator the gold answer and asks it to write a chain that reaches that answer. We show that this second step degrades the training data in a way that correctness filtering cannot catch. We run a controlled experiment that fixes the generator, the problem set, and the correctness filter, and varies only whether the chain is generated under answer-conditioning, the gold answer shown with a request to reach it. Training a strong instruction-tuned reasoning model on its own answer-conditioned chains sharply lowers its verifiable-reasoning accuracy. The loss grows with difficulty, reaching as much as about $27$ points on the hardest competition problems. The mechanism is legible in the chains themselves, which rationalize backward from the shown answer instead of deriving it, with the early final-answer statement as the measurable symptom. The harm is a property of the data rather than the generator, read off unlabeled generations before any fine-tuning, ordering the penalty across eight thinking models from four families, and transferring across teacher families. A prompt ablation localizes it to the rationalize-toward instruction rather than the answer's bare visibility. The practical takeaway is to generate answer-blind, because no correctness filter can see this damage in the data. Code is available at \url{https://github.com/js-lee-AI/answer-leakage}.
\end{abstract}

\section{Introduction}
\label{sec:intro}

\begin{figure*}[t]
\centering
\includegraphics[width=0.9\textwidth]{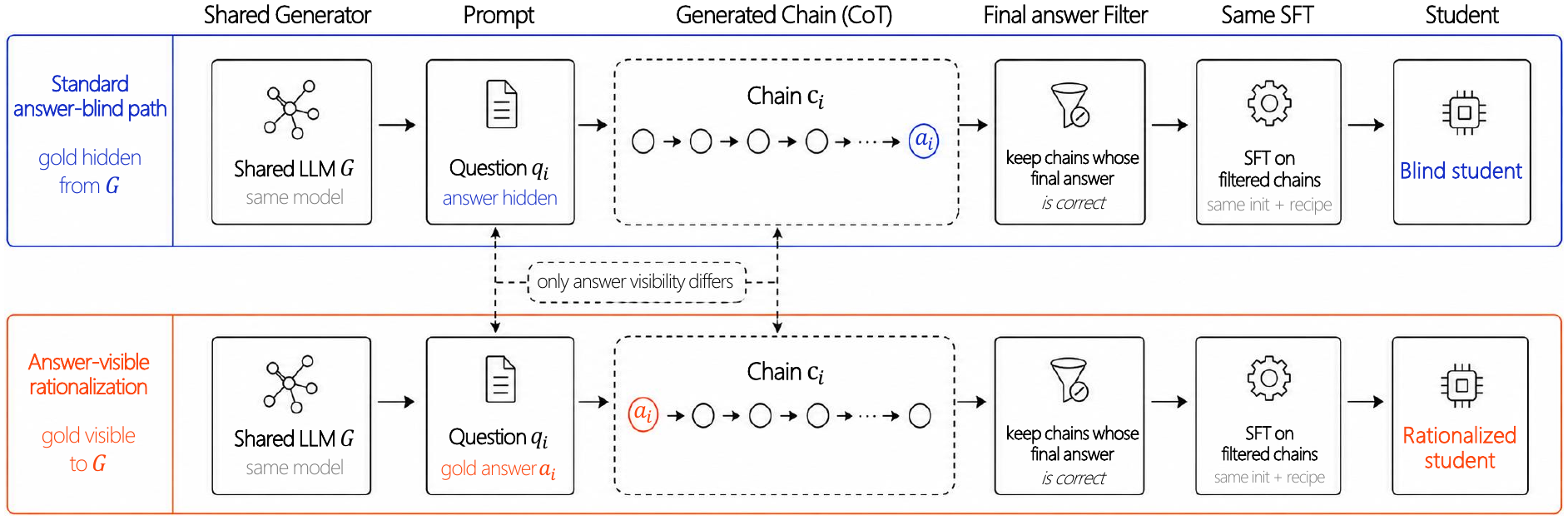}
\caption{The one-bit experiment. A shared generator produces chains for the same problems under an answer-blind path and an answer-visible rationalization path. Both arms use the same final-answer filter and the same SFT recipe, so the arms differ only in whether generation is answer-conditioned, the gold answer shown together with a request to reach it.}
\label{fig:fig1}
\end{figure*}

Large language models reason through chains of thought \citep{wei2022chain,openai2024o1card}, and much recent progress in open reasoning models comes from distilling this ability by supervised fine-tuning on the models' own chains of thought \citep{deepseekai2025deepseekr1}. A typical pipeline samples candidate chains, keeps the ones that reach the correct final answer, and fine-tunes the model on the survivors. When sampling fails to produce a correct chain, or when the available chains look noisy, a popular fix is to condition the generation on the gold answer. The self-taught reasoner method feeds the answer as a hint and asks the model to rationalize problems it could not solve \citep{zelikman2022star}, and answer-aware refinement pipelines regenerate chains directly from question and answer pairs \citep{zhang2023hindsight,yu2024metamath,li2024mugglemath,li2025start}. Chains produced this way still reach the correct answer, so the correctness filter keeps them.

The fix rests on the implicit assumption that a correct chain is a good training target no matter how it was produced \citep{yuan2023rft,singh2023beyond,xiong2025hsstar,li2026kstar}. We show that this assumption fails. A chain that reaches the right answer can still teach the wrong behavior when the model wrote it while looking at the answer, and the standard correctness filter cannot catch the problem, because the chain is correct by construction.

To isolate the effect we run what we call the one-bit experiment, summarized in Figure~\ref{fig:fig1}. Using a single model, we generate chains of thought for the same set of problems under two conditions that differ by a single intervention, whether generation is answer-conditioned, the gold answer shown in the prompt with a request to reach it. We keep only correct chains in both conditions and fine-tune with an identical recipe. Training on the answer-conditioned chains is sharply worse than training on the answer-blind ones. The drop persists under a length-matched control and a budget sweep and grows with difficulty. Its size depends on the model, destroying accuracy on an already-strong one and forfeiting part of a real answer-blind gain on one with room to improve.

Throughout, answer leakage is our informal shorthand for this recipe, the move that deployed self-improvement pipelines make when sampling fails \citep{zelikman2022star,zhang2023hindsight,yu2024metamath,li2025start}. The recipe bundles two ingredients, the answer's visibility and the instruction to rationalize toward it. Our mechanism analysis separates them, and the instruction is what carries the harm. At fixed visibility a neutral instruction recovers most of the penalty.

A chain composed toward a known answer rationalizes instead of deriving, and its symptom is stating the answer near the start. Within a single corpus, we find that these rationalized chains carry most of the harm, and that a simple filter dropping them is a partial repair. The same tendency is measurable before any fine-tuning, where it predicts the penalty across eight models from four families, transfers on average when answer-conditioned chains from one model train a smaller one. We study verifiable-reasoning distillation, domains such as math and code where a final-answer check or test execution lets the correctness filter run.

The effect is large where a derivation is required and absent where it is not. For Qwen3-8B, the answer-conditioned corpus trains a student $16.2$ points weaker on MATH-500 than the answer-blind corpus built from the same problems, and the gap grows to $27.2$ points on olympiad AIME problems. Repeating the intervention in code degrades the student as well, even though every kept chain passes its tests. On multiple-choice knowledge and science benchmarks, which ask the model to select an option rather than write a derivation, we find no penalty, so the harm is confined to tasks whose answers must be derived.

Our contributions are the following.
\begin{enumerate}
  \item We isolate the cost of answer-conditioned generation. Holding the generator, problems, correctness filter, data volume, and chain length fixed, we show that generating chains under answer-conditioning incurs the leakage penalty, we localize it by a prompt ablation to the instruction to rationalize toward the answer rather than to merely seeing it, and we replicate it on further models.
  \item We trace the mechanism. We find that a visible answer moves chains toward stating the answer early, an off-policy shift away from the generator's own search distribution, and we show that the habit transfers to the student's own outputs even with answer correctness controlled.
  \item We attribute the harm to the rationalized chains and partly repair it with a filter. In an experiment inside a single answer-conditioned corpus, we show that these chains carry most of the harm, a three-seed difference-in-differences of $16.9$ points.
  \item We predict the penalty before any training. We read a signature off unlabeled generations that orders the leakage penalty across eight models from four families and predicts a held-out family with a mean absolute error of $2.9$ points.
\end{enumerate}

These findings come with cheap safeguards. A candidate teacher can be screened before any student is trained, since the signature is measured on a small set of its unlabeled generations, and where a pipeline cannot avoid showing the gold answer, an instruction to derive first and only then state the answer restores about two-thirds of the lost accuracy. The corrective costs nothing extra, since generating answer-blind with the same correctness filter avoids the pathology and modestly beats distillation from gold data at a controlled budget.

\section{Related Work}
\label{sec:related}

\paragraph{Rationalization in self-improvement pipelines.}
A prominent line of self-improvement methods treats final-answer correctness as the quality criterion for a chain~\citep{xie2025coped}. The self-taught reasoner approach keeps the rationales that reach correct answers and rationalizes the rest by feeding the model the answer as a hint \citep{zelikman2022star}, a hindsight relabeling of failed attempts \citep{zhang2023hindsight}. The same move extends to token-level reasoning \citep{zelikman2024quietstar} and recurs in rejection-sampling and ReST-style methods \citep{yuan2023rft,gulcehre2023rest,singh2023beyond,zhang2024restmcts,luong2024reft,zhu2026intoken} and in answer-augmentation pipelines such as MetaMath \citep{yu2024metamath}.

A closely related analysis, TRICE, shows that conditioning a chain on the answer biases the learning signal and recasts the objective as latent-variable inference to correct it \citep{phan2023training}. We instead show that this bias is invisible to the standard correctness filter, causally harms supervised fine-tuning even when every chain is verified, and can be anticipated from unlabeled generations before any training. A different strategy trains a verifier on both correct and incorrect chains rather than filtering them away \citep{vstar2024}. Our results show that how a correct chain was produced still matters. To our knowledge, ours is the first controlled isolation of answer-conditioning, holding the generator, problems, correctness filter, and recipe fixed and varying only whether the generator is shown the answer.

\paragraph{Backward reasoning without leakage.}
Reverse-reasoning methods that generate backward questions or check forward and backward consistency without revealing the gold answer are leakage-free and are not implicated by our findings \citep{chen2024reverse,kazemi2023lambada,weng2023large}. We study the answer-visible regime that those methods avoid.

\paragraph{Process vs.\ outcome supervision.}
Outcome-supervised verifiers \citep{cobbe2021training}, comparisons of process and outcome feedback \citep{uesato2022solving}, and process reward models \citep{lightman2023lets,wang2024math,luo2024improve,zheng2025processbench,song2025prmbench,zhang2025lessons,wang2026outcome} all argue that final-answer correctness is an insufficient quality signal and supervise or rerank individual steps. We show a complementary failure on the data-generation side, where chains that are correct and step-wise plausible can still harm the student because of how they were generated.

\paragraph{Faithfulness of chain-of-thought.}
Chains of thought can be unfaithful or post-hoc. They are unfaithful when a model exploits a hint injected at inference time without verbalizing it \citep{turpin2023language,chen2025reasoning,tutek2025unlearning,xu2024preemptive}, and post-hoc when the stated reasoning does not drive the final answer \citep{lanham2023measuring,paul2024making}. We show the training-time analogue, where hint-conditioned targets transfer the post-hoc habit to the student, producing a measurable shift in answer position and an accuracy cost.

\paragraph{Data selection and on-policy distillation.}
Data-selection work argues that fewer and better examples can outperform larger noisy sets \citep{zhou2023lima}, but our harmful examples already pass the correctness criterion, so the failure mode is distinct from low quality. On-policy distillation reports advantages for self-generated data under a soft-KL objective \citep{agarwal2024onpolicy,sang2026crisp}, which is consistent with our answer-blind result, but that setting changes the generator and the objective at the same time and never isolates answer visibility.

\paragraph{Concurrent leakage filters.}
Concurrent work has begun to patch leakage symptoms inside refinement pipelines, for example gating hindsight-hint repairs with an entropy filter \citep{zhang2026heal,zhang2026dasd,li2026raft}, but such mitigations do not isolate what is being filtered or why. Our one-bit experiment supplies the missing causal quantity, the accuracy cost of answer-conditioned generation in an otherwise identical pipeline, together with a test isolating which chains carry the penalty inside one corpus and a predictor read off unlabeled generations before any fine-tuning.

\section{Isolating Answer Leakage}
\label{sec:g1}

\begin{table}[t]
\centering
\small
\setlength{\tabcolsep}{5pt}
\renewcommand{\arraystretch}{1.15}
\resizebox{0.9\columnwidth}{!}{%
\begin{tabular}{@{}l c c r@{}}
\toprule
Setting / Benchmark & Blind\,$\uparrow$ & Leaked\,$\uparrow$ & Penalty\,$\downarrow$ \\
\midrule
\multicolumn{4}{@{}l}{\textbf{\emph{(A) Instruction ablation at fixed visibility}}} \\
Rationalize toward answer & \cellcolor{blue!6}95.1 & 78.9 & \textcolor{red}{$16.2$} \\
Derive-first              & \cellcolor{blue!6}95.1 & 90.7 & \textcolor{red}{$4.3$} \\
Final-check               & \cellcolor{blue!6}95.1 & 93.7 & \textcolor{red}{$1.3$} \\
Out-of-band               & \cellcolor{blue!6}95.1 & 94.3 & \textcolor{red}{$0.7$} \\
\midrule
\multicolumn{4}{@{}l}{\textbf{\emph{(B) Within-corpus chain-population carrier}}} \\
\textsc{Leaked-Early} & \cellcolor{blue!6}94.0 & 64.2 & \textcolor{red}{$29.8$} \\
\textsc{Leaked-Late}  & \cellcolor{blue!6}94.8 & 86.2 & \textcolor{red}{$8.6$} \\
\addlinespace
Difference-in-differences & & & $+16.9$ \\
\midrule
\multicolumn{4}{@{}l}{\textbf{\emph{(C) Across math benchmarks, difficulty-ordered}}} \\
GSM8K \textit{\footnotesize (grade school)}       & \cellcolor{blue!6}95.4 & 90.5 & \textcolor{red}{$4.9$} \\
MATH-500 \textit{\footnotesize (competition)}     & \cellcolor{blue!6}95.1 & 78.9 & \textcolor{red}{$\mathbf{16.2}$} \\
\quad Level 1              & \cellcolor{blue!6}95.3 & 95.3 & $0.0$ \\
\quad Level 2              & \cellcolor{blue!6}98.1 & 84.1 & \textcolor{red}{$14.1$} \\
\quad Level 3              & \cellcolor{blue!6}98.4 & 83.5 & \textcolor{red}{$14.9$} \\
\quad Level 4              & \cellcolor{blue!6}95.6 & 74.2 & \textcolor{red}{$21.4$} \\
\quad Level 5              & \cellcolor{blue!6}89.8 & 70.9 & \textcolor{red}{$18.9$} \\
Minerva \textit{\footnotesize (graduate STEM)}    & \cellcolor{blue!6}48.7 & 38.5 & \textcolor{red}{$10.2$} \\
AIME 2024--2025 \textit{\footnotesize (olympiad)} & \cellcolor{blue!6}71.7 & 44.4 & \textcolor{red}{$27.2$} \\
\bottomrule
\end{tabular}
}%
\caption{Answer-leakage penalty on Qwen3-8B, MATH-500 accuracy unless a benchmark is named, with Penalty the Blind minus Leaked gap of Eq.~\eqref{eq:penalty}. The tinted Blind column is the recommended answer-blind arm. (A) varies the generation instruction at fixed answer visibility. (B) is the within-corpus two-by-two carrier design. (C) spans four difficulty-ordered math benchmarks.}
\label{tab:main}
\end{table}

\begin{figure*}[t]
\centering
\includegraphics[width=0.9\textwidth]{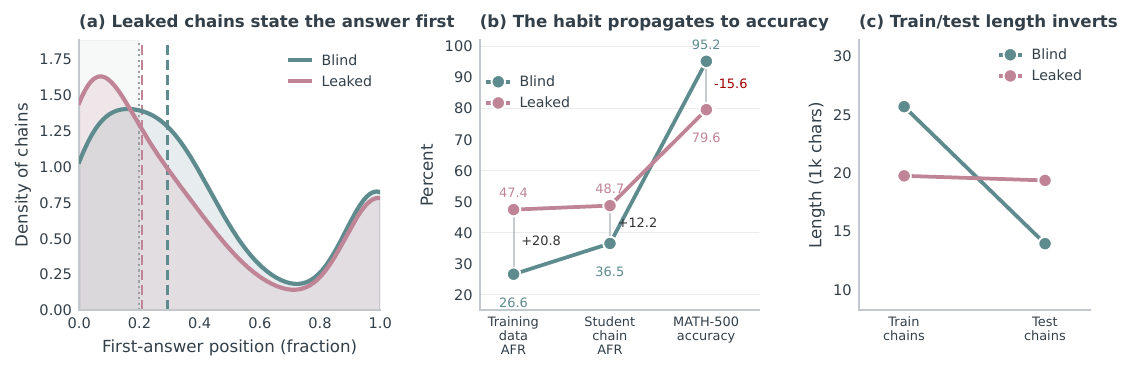}
\caption{Answer-position and length behavior of the answer-blind and answer-leaked arms in training chains and student outputs. \textbf{(a)} Distribution of the first gold-answer position within the think block for the answer-blind chains, generated with the answer never shown, and the answer-leaked chains, generated with the gold answer shown. \textbf{(b)} Answer-first rate ($\afr$, Eq.~\eqref{eq:afr}) in the training corpus and in the trained students' own answer-blind generations, for the blind and leaked arms. \textbf{(c)} Mean think-block length of the blind and leaked arms in the training chains and in the students' own test generations.}
\label{fig:mech}
\end{figure*}

Unless noted otherwise, every trained model is a full-parameter supervised fine-tune of Qwen3-8B~\citep{qwen3techreport} under one fixed recipe, on training problems from the math split of the Nemotron-Cascade-SFT-Stage-2 corpus \citep{wang2025nemotroncascade}, disjoint from all evaluation sets. We evaluate with vLLM \citep{kwon2023efficient} in thinking mode on MATH-500 \citep{hendrycks2021measuring} with a $32{,}768$-token budget and symbolic answer matching. Under this harness the Qwen3-8B base model scores $95.2$. Appendix~\ref{app:details} gives the full recipe, sampling, and harness details. The comparison is built on a fixed set of $935$ math problems, each carrying a gold answer $a(q)$. Holding all of this fixed, the lone intervention is answer-conditioning.

Let $M$ be the generator. For each problem we build two prompts. The \nohint{} prompt (answer-blind) shows only the problem and asks the model to derive the answer. The \hint{} prompt (answer-leaked) appends $a(q)$ and asks for a reasoning trace that arrives at it. From $M$ we sample chains under each prompt, keep only chains whose final boxed answer is correct, and fine-tune two students on the resulting (question $\rightarrow$ chain) pairs under one identical recipe, an answer-blind student $S_{\mathrm{blind}}$ and an answer-leaked student $S_{\mathrm{leaked}}$. The reported leakage penalty is the accuracy lost by generating the chain under answer-conditioning,
\begin{equation}
\Delta = \operatorname{acc}(S_{\mathrm{blind}}) - \operatorname{acc}(S_{\mathrm{leaked}}).
\label{eq:penalty}
\end{equation}
The generator, problem set, correctness filter, recipe, and the matched examples are held fixed, so the two students differ only in answer-conditioning and in the chain-length shift it induces, which we control with a length-matched arm below.

Table~\ref{tab:main} shows that generating under answer-conditioning costs $16.2$ points on MATH-500 as a three-seed mean. Because every other factor is held fixed, this gap is the leakage penalty $\Delta$ of Eq.~\eqref{eq:penalty}. The \nohint{} student sits at the base model within seed noise ($95.1$ against $95.2$), so answer-blind self-distillation preserves performance and essentially the whole gap is leakage damage. Table~\ref{tab:main}(C) evaluates the same fine-tunes on four independent math benchmarks from grade-school GSM8K to olympiad AIME, and the penalty holds across all four. Within MATH-500 it grows with problem difficulty, from near zero on the easiest problems to over twenty points on the hardest levels. On the pooled Level~4--5 band the penalty is $20.1$ points with real headroom in the blind arm, with the band-level and seed-level numbers in Appendix~\ref{app:g1levels}. Appendix~\ref{app:details} reports the seed-level scores and Appendix~\ref{app:stats} the full inferential statistics.

\paragraph{The penalty is content, not length.}
Appendix~\ref{app:details} reports a length-matched control that rules out length as the explanation. To check that length is not the cause, we truncate each \emph{blind} chain, which runs longer, to its matched \emph{leaked} chain's length, preserving genuine answer-blind reasoning and the correct answer, and fine-tune on it. This length-matched arm scores $95.0$, far above \hint{}, so the penalty is carried by the answer-conditioned \emph{content}, not the length.

\begin{figure}[t]
\centering
\includegraphics[width=0.9\linewidth]{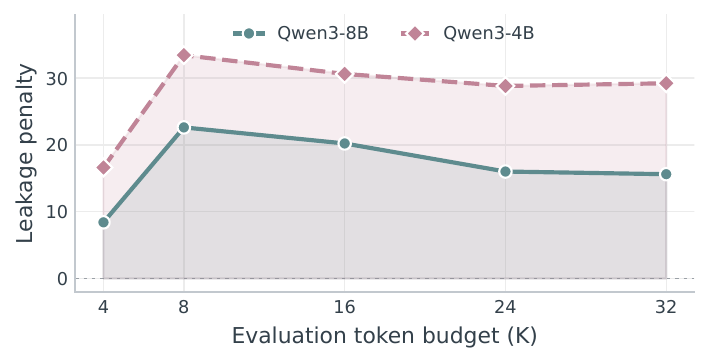}
\caption{Leakage penalty against the evaluation token budget, scoring a problem correct only when its boxed answer completes within the budget.}
\label{fig:budget}
\end{figure}

\section{Mechanism Analysis}
\label{sec:anatomy}

Answer-conditioning pushes the model to state the final answer early and narrate toward it rather than derive it, a shift measurable in unlabeled generations before any fine-tuning. We read it off the matched pairs through answer position and the student's test-time behavior, account for it as an off-policy shift in the generation distribution, and then ask whether the same shift identifies which chains inside a single corpus carry the harm.

\subsection{How the leakage penalty manifests}

\paragraph{A visible answer makes the model state the answer early.}
This reading rests on a single chain-level property. A chain is \emph{answer-first} if the gold final answer is stated within the first $20\%$ of the think block, and \emph{derivation-first} otherwise. Writing $\mathrm{pos}(c)\in[0,1]$ for the relative position of the first gold-answer mention inside the reasoning of a chain $c$, the \emph{answer-first rate} of a corpus $C$ is
\begin{equation}
\afr(C) = \frac{1}{|C|}\sum_{c \in C}\mathbf{1}\!\left[\mathrm{pos}(c) < 0.2\right].
\label{eq:afr}
\end{equation}
A high $\afr$ is the measurable symptom of rationalizing toward a known answer rather than deriving it. Figure~\ref{fig:mech}a shows the leaked chains landing in the answer-first zone far more often than blind, with the whole position distribution shifted toward the start of the think block. On the matched pairs this is a $20.8$-point rise in $\afr$ at comparable answer coverage, detailed in Appendix~\ref{app:anatomy}.

Table~\ref{tab:main}A holds the answer visible across all arms and varies only the instruction. The penalty falls with the answer-first rate to under a point, so the harm tracks the instruction, not the answer's mere visibility. The derive-first instruction shifts the answer-first rate to $36.0\%$, between the leaked $47.4\%$ and the blind $26.6\%$, and recovers about two-thirds of the penalty. The two neutral arms still land about a point below blind, a residual cost of bare visibility set against the fifteen the instruction carries.

\begin{table}[t]
\centering
\small
\setlength{\tabcolsep}{6pt}
\renewcommand{\arraystretch}{1.05}
\resizebox{0.75\columnwidth}{!}{%
\begin{tabular}{lcc}
\toprule
& \multicolumn{2}{c}{Answer-first rate $\afr$} \\
\cmidrule(lr){2-3}
Student outputs & All & Correct \\
\midrule
Base \textit{\footnotesize (no SFT)}             & 38.7 & 37.1 \\
\nohint{} \textit{\footnotesize (answer blind)}  & \cellcolor{blue!6}36.5 & \cellcolor{blue!6}35.8 \\
\hint{} \textit{\footnotesize (answer leaked)}   & 48.7 & 45.2 \\
\addlinespace
$\afr$ leaked $-$ blind  & $+12.2$ & $+9.4$ \\
\bottomrule
\end{tabular}
}%
\caption{Answer-first rate ($\afr$, Eq.~\eqref{eq:afr}) of each student's own MATH-500 generations under the answer-blind solve prompt. The tinted \nohint{} row is the recommended answer-blind arm.}
\label{tab:mech}
\end{table}

\paragraph{Two deficits underlie the penalty.}
Figure~\ref{fig:budget} shows the penalty growing as the evaluation budget shrinks, the mark of a termination deficit, and the damage splits into two measurable components. The \emph{derivation deficit} is error when the student does finish, on problems where both students produce an extractable answer the leaked one is still $9.5$ points worse on Qwen3-8B and $21.1$ on Qwen3-4B. The \emph{termination deficit} is failure to finish at all, the leaked student producing no extractable answer within the budget on $7.4$ of every hundred problems against $0.4$ for blind on Qwen3-8B, and $12.6$ against $1.8$ on Qwen3-4B. We trace this trained-in pathology to a train/test length inversion.

\paragraph{Leaked chains invert the train/test length relationship.}
Figure~\ref{fig:mech}c shows the starkest behavioral fingerprint, a train/test length inversion. At \emph{training} time the leaked chains are \emph{shorter} because seeing the answer compresses the reasoning, yet at \emph{test} time the leaked-trained model generates \emph{longer} chains than the blind-trained one while scoring worse. Trained on short answer-aware chains, it never learned to derive an unknown answer and generates at length without converging, the termination deficit above.

\subsection{The habit is off-policy rationalization}

\paragraph{The answer-first habit transfers to the student's own outputs.}
Table~\ref{tab:mech} shows that the tendency is not confined to the training corpus, reappearing in each student's own MATH-500 generations where no answer is shown, so the leaked student has internalized the habit. Scoring $\afr$ (Eq.~\eqref{eq:afr}) on each student's evaluation chains against each chain's own boxed answer, the leaked student is answer-first more often than blind, which sits with the base. A correctness control on the $387$ problems all three students solve preserves a $9.4$-point gap, and the median first-answer position stays earlier for the leaked student. The transfer is partial, because leaked chains run about $29\%$ longer, as Appendix~\ref{app:mech} details.

\paragraph{Answer-conditioned chains are off-policy rationalizations, not search.}
We read these chains as \emph{rationalizations}, coherent narratives toward a known conclusion that satisfy outcome filters while encoding a search-free policy whose symptom is the high answer-first rate. The answer-blind generator samples from $P(c \mid q)$, the search distribution a deployed solver must follow because the answer is unknown at test time. Answer-conditioning instead samples from the lower-entropy $P(c \mid q, a(q))$ and collapses that search to post-hoc justification of a known target. Hard supervised fine-tuning is on-policy only with $P(c \mid q)$ targets. Targets drawn from $P(c \mid q, a(q))$ are therefore off-policy and teach a distribution the student cannot occupy once the answer is gone, exactly the early-answer narration we observe. Two questions follow, whether the damage travels with the rationalized chains inside a single corpus and whether $\Delta\afr$ flags the harmed models before training.

\subsection{Isolating the rationalized chains}
\label{sec:carrier}

The first open question is whether the harm concentrates in the rationalized chains. Inside the single leaked corpus the mechanism account expects it to, with a chain-level filter that drops those chains acting as a drop-in fix. We test this causally.

\paragraph{A within-corpus split isolates the rationalized chains.}
We split the $935$ leaked chains by whether the answer is stated within the first $20\%$ of the think block, giving \textsc{Leaked-Early} ($438$ chains) and \textsc{Leaked-Late} ($497$), fine-tune separately on each with the same recipe, and read the four arms in Table~\ref{tab:main}(B).

The split conditions on a leaked-chain property, so it could track problem difficulty, and the blind twins of the two subsets confirm that it partly does, as Appendix~\ref{app:twobytwo} quantifies. We therefore add two control arms on the blind chains of the same problems, \textsc{Blind-PE} and \textsc{Blind-PL}, turning the comparison into a difference-in-differences (DiD) that nets out problem-subset effects. A length account predicts the opposite ordering here, because the rationalized chains are the longer ones, $21.4$K against $18.3$K characters. Appendix~\ref{app:twobytwo} confirms that the split is robust to the answer matcher, the position threshold, and the handling of chains that never state the answer.

\paragraph{Primary endpoint.}
Reusing the $0.2$ answer-first threshold from the signature above, the primary endpoint is the difference-in-differences
\begin{equation}
\begin{aligned}
\mathrm{DiD} = {}&(\textsc{Leaked-Late} - \textsc{Leaked-Early}) \\
&- (\textsc{Blind-PL} - \textsc{Blind-PE}),
\end{aligned}
\label{eq:did}
\end{equation}
evaluated over the $500$ evaluation problems. Exact thresholds in Appendix~\ref{app:twobytwo} mark three outcomes, a full carrier when \textsc{Leaked-Late} recovers to the blind arms, a partial carrier when it sits well above \textsc{Leaked-Early} but still below blind, and no carrier when it is not decisively positive.

\paragraph{The rationalized arm carries most of the harm.}
Table~\ref{tab:main}(B) reports the four arms and the primary endpoint. The model trained on rationalized leaked chains collapses while the derivation-first arm largely holds. The two blind controls do not differ, so the easy-answer problems train perfectly well once the chains are blind. The three-seed mean of the endpoint is $+16.9$ (seed range $12$ to $21$).\footnote{A single-seed run gives $+21.2$.}

Against their own blind twins the rationalized arm loses about $3.5\times$ as much as the derivation-first arm. The rationalized chains therefore carry most but not all of the harm, and a chain-level filter is a cheap partial repair that leaves a residue only blind generation closes. We scope the causal claim to the chain population \emph{marked by} the signature and sharpen it to the single answer-stating line when we probe generalization. Appendix~\ref{app:stats} reports the full interval analysis and Appendix~\ref{app:twobytwo} the robustness of the split.

\begin{table}[t]
\centering
\small
\setlength{\tabcolsep}{3pt}
\renewcommand{\arraystretch}{1.1}
\resizebox{0.9\columnwidth}{!}{%
\begin{tabular}{lccccc}
\toprule
& \multicolumn{2}{c}{\textit{Pre-FT signature}} & \multicolumn{3}{c}{\textit{Post-FT outcome}} \\
\cmidrule(lr){2-3}\cmidrule(lr){4-6}
Model & Base\,$\uparrow$ & $\Delta\afr$ & Blind\,$\uparrow$ & Leaked\,$\uparrow$ & Penalty\,$\downarrow$ \\
\midrule
\multicolumn{6}{@{}l}{\textbf{\emph{Held-in models}}} \\
Qwen3-8B            & 95.2 & $+20.8$ & 95.1 & 78.9 & \textcolor{red}{$16.2$} \\
Qwen3-4B            & 96.0 & $+26.6$ & 94.0 & 64.8 & \textcolor{red}{$29.2$} \\
GLM-Z1-9B           & 94.4 & $+11.1$ & 94.8 & 82.6 & \textcolor{red}{$12.2$} \\
DS-Qwen-7B          & 79.6 & $+11.0$ & 89.7 & 82.4 & \textcolor{red}{$7.3$} \\
DS-Llama-8B         & 77.8 & $-0.2$  & 65.8 & 64.6 & \textcolor{red}{$1.2$} \\
\addlinespace
\multicolumn{6}{@{}l}{\textbf{\emph{Held-out models}}} \\
Qwen3-1.7B  & 92.6 & $+26.2$ & 89.6 & 59.8 & \textcolor{red}{$\mathbf{29.8}$} \\
DS-Qwen-14B & 88.2 & $+9.3$  & 92.2 & 83.4 & \textcolor{red}{$\mathbf{8.8}$} \\
Nemotron-8B & 95.4 & $+17.9$ & 92.2 & 74.6 & \textcolor{red}{$\mathbf{17.6}$} \\
\bottomrule
\end{tabular}
}%
\caption{The answer-first signature across eight thinking models from four families, each self-distilled blind against leaked on MATH-500. Base is un-fine-tuned accuracy, $\Delta\afr$ the pre-fine-tuning signature of Eq.~\eqref{eq:afr}. The Qwen3-8B and DS-Qwen-7B rows are three-seed means, the other six single training runs.}
\label{tab:signature}
\end{table}

\section{Anticipating the Penalty Across Models}
\label{sec:predict}

Whether answer conditioning harms a model is neither universal nor random. It is anticipated, before any fine-tuning, by $\Delta\afr = \afr_{\text{leaked}} - \afr_{\text{blind}}$, which measures how much more often the model states the gold answer early when it can see it. Computing it requires only a small unlabeled sample of generations from the candidate teacher, with no labels beyond the training answers already in hand.

\begin{figure}[t]
\centering
\includegraphics[width=0.9\linewidth]{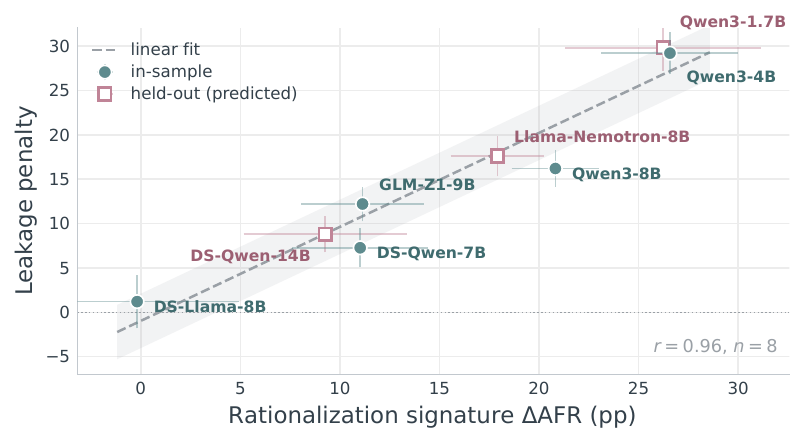}
\caption{The $\Delta\afr$ signature against the answer-leakage penalty across eight thinking models from four families. The dashed line is the linear fit and the shaded band its dispersion. Three points are out-of-sample, and labels are colored by in-sample against held-out. Error bars show sampling variability on both axes.}
\label{fig:signature}
\end{figure}

\paragraph{The answer-first signature orders the penalty across four models.}
Table~\ref{tab:signature} shows that the larger a model's $\Delta\afr$, the larger its penalty. The two track together across the first four thinking models at $r=0.96$ on a small sample. Figure~\ref{fig:signature} plots the same relation.

The penalties run from a substantial cost on Qwen3-8B through an intermediate cost on DeepSeek-R1-Distill-Qwen-7B to a negligible one on DeepSeek-R1-Distill-Llama-8B. The penalty therefore scales continuously with the signature rather than switching on for some models and off for others. One further thinking model, Phi-4-reasoning, cannot be scored for measurement reasons, so we report the fit as an interpolation over the measured range, with the details in Appendix~\ref{app:stats}.

\paragraph{The fit predicts two held-out models before training them.}
\begin{table}[t]
\centering
\small
\setlength{\tabcolsep}{6pt}
\renewcommand{\arraystretch}{1.12}
\begin{tabular}{@{}lcc@{}}
\toprule
& \multicolumn{2}{c}{Leakage Penalty} \\
\cmidrule(l){2-3}
Held-out model & Predicted & Observed \\
\midrule
Qwen3-1.7B & 24.4 & 29.8 \\
DS-R1-Distill-Qwen-14B & 7.5 & 8.8 \\
Llama-Nemotron-8B \textit{\footnotesize (third family)} & 17.4 & 17.6 \\
GLM-Z1-9B & 10.1 & 12.2 \\
\bottomrule
\end{tabular}
\caption{Held-out leakage penalties predicted from the answer-first signature against the penalties later observed. Each prediction was made before that model's fine-tunes were evaluated. Llama-Nemotron-8B is an independent third family.}
\label{tab:predict}
\end{table}

Table~\ref{tab:signature} marks the held-out models with bold observed values. A completely different model family, NVIDIA's Llama-Nemotron-8B~\citep{bercovich2025llamanemotron} independent of both Qwen and DeepSeek, is predicted to lose $17.4$ points and loses $17.6$. Table~\ref{tab:predict} collects the four held-out predictions against their observed penalties, every one landing within its predicted band, with the fits in Appendix~\ref{app:beyond}. Retraining the two held-out predictor arms on a four-times expanded matched set of $807$ pairs, whose $\Delta\afr=+25.3$ stays consistent, gives blind $90.6$ against leaked $65.8$, a penalty of $+24.8$ within $0.4$ points of the fit's prediction. Across all eight models the signature explains the penalty closely at $r=0.960$, and no single point carries it, with the intervals reported in Appendix~\ref{app:stats}.

\paragraph{The answer-first signature survives the capability and family confounds.}
Appendix~\ref{app:stats} reports that the signature survives two covariates that could explain the relation away. Base accuracy correlates with the penalty at $0.76$, yet controlling for it the penalty still follows the signature at a partial correlation of $0.91$, and holding out an entire family and predicting it from the other three gives a mean absolute error of $2.9$ points. More fundamentally, the carrier experiment of Section~\ref{sec:carrier} tests the same mechanism \emph{within} a single model and corpus, where neither confound can exist.

\paragraph{The leaked arm loses in both regimes, differing only in magnitude.}
Table~\ref{tab:signature} splits the eight models by base accuracy into a \emph{saturated} regime, where SFT adds little and leakage destroys an already-strong model, and a \emph{headroom} regime, where blind SFT gains but leakage forfeits part of that gain, yet blind never loses to leaked. On the headroom base DeepSeek-R1-Distill-Qwen-7B blind self-distillation reaches $89.7$ over three seeds, clearing the $79.6$ base, while the leaked arm forfeits part of that gain at $82.4$, a $+7.3$ penalty positive in every seed, with the seed-level scores in Appendix~\ref{app:beyond}.

\section{Generalization}
\label{sec:beyond}

\begin{table}[t]
\centering
\small
\setlength{\tabcolsep}{6pt}
\renewcommand{\arraystretch}{1.1}
\begin{tabular}{@{}l c c r@{}}
\toprule
Setting & Blind\,$\uparrow$ & Leaked\,$\uparrow$ & Penalty\,$\downarrow$ \\
\midrule
\multicolumn{4}{@{}l}{\textbf{\emph{(A) Code domain, Qwen3-8B student}}} \\
MBPP+ \textit{\footnotesize (pass rate)}      & \cellcolor{blue!6}68.3 & 58.8 & \textcolor{red}{$9.5$} \\
HumanEval+ \textit{\footnotesize (pass rate)} & \cellcolor{blue!6}72.8 & 63.6 & \textcolor{red}{$9.2$} \\
\midrule
\multicolumn{4}{@{}l}{\textbf{\emph{(B) Smaller student}}} \\
Cross-student \textit{\footnotesize (1.7B from 8B)} & \cellcolor{blue!6}85.7 & 78.9 & \textcolor{red}{$6.9$} \\
\midrule
\multicolumn{4}{@{}l}{\textbf{\emph{(C) Cross-family teacher, Qwen3-8B student}}} \\
Nemotron & \cellcolor{blue!6}94.0 & 79.2 & \textcolor{red}{$14.8$} \\
GLM      & \cellcolor{blue!6}93.0 & 82.8 & \textcolor{red}{$10.2$} \\
\bottomrule
\end{tabular}
\caption{The one-bit intervention carried to the code domain, a smaller student, and two further teacher families with a Qwen3-8B student. Columns are Blind and Leaked accuracy and the red Penalty is Blind $-$ Leaked, MATH-500 accuracy for the math rows and pass rate for the code rows.}
\label{tab:transfer}
\end{table}

Beyond the four math benchmarks of Table~\ref{tab:main}(C), the one-bit experiment still fixes one student that also serves as generator, one recipe, and one domain, and relaxing each in turn leaves the effect intact. Table~\ref{tab:transfer} collects those relaxations, where the penalty holds with the same sign across every generative setting and vanishes only where the task needs no multi-step derivation, the boundary the mechanism predicts.

\paragraph{The damage travels with the data, not the student.}
Appendix~\ref{app:beyond} reports that the harm follows the corpus when a strong teacher's chains train a different, smaller student. Fine-tuning Qwen3-1.7B on the Qwen3-8B corpora over the same $935$ problems still leaves the leaked arm about $7$ points below blind on average across three seeds, though the instability reflects small-student fine-tuning variance rather than the data. STaR-style rescue~\citep{zelikman2022star} that leaks the answer only on failed problems still costs a real $4.0$-point hit, as Table~\ref{tab:star} reports, so even verified rescued chains are negative-value data.

\paragraph{Two unrelated teacher families carry the same penalty.}
Table~\ref{tab:transfer} reports the penalties that chains from two further teacher families impose on an unrelated Qwen3-8B student. When a Llama-3.1-Nemotron-Nano-8B teacher, a from-scratch Llama lineage distinct from Qwen and DeepSeek, supplies the chains, its leaked corpus costs the student a $14.8$-point penalty. A GLM-Z1-9B teacher, an unrelated fourth family, costs a $10.2$-point penalty. In the predictor of Table~\ref{tab:predict} each model both wrote and learned its own chains, so teacher and student were confounded, whereas here the teacher hands its corpus to a model of a separate family, which isolates the corpus. A length-matched control on the GLM transfer rules out length, as Appendix~\ref{app:beyond} details. Whether the student is the teacher itself, a smaller sibling, or a model from a separate family, one corpus of answer-conditioned chains lowers the accuracy of every student it trains.

\paragraph{A single answer-stating line carries most of the harm.}
The carrier experiment of Section~\ref{sec:carrier} marks the answer-first \emph{population}, yet correlated properties could still ride along, so we excise from each \textsc{Leaked-Early} chain the single line where the gold answer is first stated, a median cut of $302$ characters or about $1.4\%$ of the chain, and retrain the otherwise-identical $438$ chains. The excised arm scores $88.8$, recovering $24.6$ of the $29.8$ points that separate \textsc{Leaked-Early} from its blind twin, or $83\%$ of the gap, and surpasses even the derivation-first leaked arm, so the early answer statement \emph{itself} is the dominant carrier.

\paragraph{The penalty extends past boxed-answer math.}
Table~\ref{tab:transfer} shows that the penalty survives in code, where the ``answer'' is the reference solution and ``correct'' means passing the tests. On MBPP+ the penalty is about $9$ to $10$ points over three seeds and HumanEval+ echoes it, so answer-conditioned chains that pass tests still harm the student. The chain-level signature transfers as well, leaked code chains echoing a verbatim fragment of the reference solution early in their reasoning $13.0\%$ of the time against $1.3\%$ for blind chains. A length-matched control rules out length for the code domain, as Appendix~\ref{app:beyond} details.

The penalty does not survive onto multiple-choice knowledge and science benchmarks, where little derivation is needed. On MMLU and GPQA-Diamond the blind and leaked students sit together near the base model as three-seed means, the differences within noise, with the values in Appendix~\ref{app:beyond}. The boundary is the one the mechanism predicts, since these tasks select an option rather than generate a derivation.

\section{Conclusion}
\label{sec:conclusion}

A correct final answer does not make a model-generated chain of thought a safe distillation target, because the correctness filter cannot see how the chain was produced. Varying only whether generation is answer-conditioned in an otherwise identical pipeline costs $16.2$ MATH-500 points. The damage rides on rationalized chains that state the known answer early instead of deriving it, and the carrier experiment attributes most of the harm to exactly those chains. The same habit is legible before any training, since $\Delta\afr$ read off a small sample of unlabeled generations orders the penalty across eight models from four families, so a candidate teacher can be screened before any fine-tuning compute is spent.

Where the answer must be shown, a derive-first instruction recovers about two-thirds of the penalty while keeping every example. Generating answer-blind avoids the pathology outright, holding base accuracy within seed noise at $95.1$ against a $95.2$ base while beating gold-data SFT at $90.8$ under a matched budget. Verifiable-reasoning distillation pipelines should therefore audit how their chains were generated, not only whether the chains end at the right answer.

\appendix

\section{Limitations}
\label{sec:limitations}

\paragraph{Scope and coverage.}
The one-bit experiment uses $935$ matched math problems on one architecture family at a time, with the content-versus-length attribution on MATH-500 and the cross-model fit small-sample at eight models. The single-seed rows of Table~\ref{tab:transfer} support a consistent sign rather than an exact magnitude, and the code domain and the student-output transfer of Section~\ref{sec:anatomy} are not length-controlled.

\section{Seed-Level Difficulty and AIME Numbers}
\label{app:g1levels}
Table~\ref{tab:aime} reports the seed-level AIME numbers behind the three-seed mean in Table~\ref{tab:main}(C). The pooled hardest MATH-500 band (Level~4--5, $262$ problems) gives blind $92.6$ against leaked $72.5$, a $20.1$-point penalty, where the blind arm has real headroom rather than sitting at ceiling, with the seed dispersions in Appendix~\ref{app:stats}.

\begin{table}[t]
\centering
\small
\setlength{\tabcolsep}{6pt}
\renewcommand{\arraystretch}{1.15}
\begin{tabular}{lccc}
\toprule
AIME 2024--2025 & Blind & Leaked & Penalty \\
\midrule
Seed $42$   & 75.0 & 48.3 & 26.7 \\
Seed $7777$ & 65.0 & 38.3 & 26.7 \\
Seed $123$  & 75.0 & 46.7 & 28.3 \\
\midrule
Mean        & 71.7 & 44.4 & 27.2 \\
\bottomrule
\end{tabular}
\caption{Seed-level AIME 2024--2025 accuracy of the headline blind and leaked Qwen3-8B fine-tunes on $60$ off-distribution competition problems, with Penalty the Blind minus Leaked difference and the bottom row the three-seed mean.}
\label{tab:aime}
\end{table}

\section{Answer-Position Statistics}
\label{app:anatomy}
Table~\ref{tab:anatomy} reports answer-position statistics for the matched pairs of Section~\ref{sec:anatomy}, as relative positions inside each chain, the faithful unit for a rationalization claim, and as answer-first rates at absolute character cutoffs. The leaked chains state the gold answer earlier on every measure, so the leaked-against-blind ordering is the same under a relative or an absolute window. The leaked chain states its answer earlier than its blind twin on $75$ of every hundred problems by absolute position, and $57$ of every hundred think blocks are shorter than the $20\%$-of-budget window in full, so an absolute-budget threshold cannot localize an early statement.

\begin{table}[t]
\centering
\small
\setlength{\tabcolsep}{5pt}
\renewcommand{\arraystretch}{1.15}
\begin{tabular}{lcc}
\toprule
 & Blind & Leaked \\
\midrule
\multicolumn{3}{l}{\emph{(a) Position and length on the $935$ matched chains}} \\
Median first-answer position          & 0.587 & 0.297 \\
Answer within first $5\%$             & 13.3 & 41.9 \\
Median think-block length (K chars)   & 24.7 & 18.2 \\
Median first mention (K chars)        & 12.2 & 4.5 \\
Answer coverage (of $935$)            & $\geq 918$ & $\geq 918$ \\
\midrule
\multicolumn{3}{l}{\emph{(b) Answer-first rate by absolute character prefix}} \\
Within $1$K chars & 12.2 & 42.0 \\
Within $2$K chars & 17.1 & 44.8 \\
Within $4$K chars & 25.8 & 48.9 \\
Within $8$K chars & 38.3 & 59.0 \\
\bottomrule
\end{tabular}
\caption{Answer-position statistics on the $935$ matched chain pairs of the mechanism analysis. Positions in (a) are relative fractions inside each chain's own reasoning, and (b) reports answer-first rates at absolute character prefixes.}
\label{tab:anatomy}
\end{table}

\section{Student-Output Answer-First Rate}
\label{app:mech}
Table~\ref{tab:mechfull} expands the student-output transfer of Section~\ref{sec:anatomy}, scoring $\afr$ (Eq.~\eqref{eq:afr}) on each student's MATH-500 evaluation chains under the standard answer-blind solve prompt against each chain's own boxed answer. Coverage below $500$ comes from chains that lack a parseable boxed answer or a complete think block. The leaked student states its answer earlier even on the all-correct subset that matches difficulty across the three students.

A length caveat applies, since leaked chains average $17{,}882$ characters against $13{,}884$ for blind. Under an absolute-character window the matched-difficulty gap narrows to between $2.3$ and $4.4$ points, though the median first-answer position stays earlier for the leaked student at $0.238$ against $0.287$.

Across its $62$ incorrect chains the leaked student is answer-first $74.2\%$ of the time against $44.7\%$ on its correct chains, so the both-correct control is the trustworthy comparison. Across the three training seeds $\{42,123,7777\}$ the whole-set gap is $12.2$, $15.8$, and $11.4$ points and the two-student both-correct gap is $9.8$, $14.1$, and $8.1$, positive in every seed, so the transfer is robust in sign.

\begin{table}[t]
\centering
\small
\setlength{\tabcolsep}{3.5pt}
\renewcommand{\arraystretch}{1.15}
\begin{tabular}{lcccc}
\toprule
Subset & Base & Blind & Leaked & $\Delta$ \\
\midrule
Whole set                       & 38.7 & 36.5 & 48.7 & $+12.2$ \\
All correct ($387$) & 37.1 & 35.8 & 45.2 & $+9.4$ \\
Both correct, 2 students ($390$) & $-$ & 35.6 & 45.4 & $+9.8$ \\
\bottomrule
\end{tabular}
\caption{Student-output answer-first rate ($\afr$, Eq.~\eqref{eq:afr}) on each student's own MATH-500 evaluation chains under the answer-blind solve prompt, by subset, where $\Delta$ is leaked minus blind. Covered chains number $189/488$ for base, $181/496$ for blind, and $223/458$ for leaked on the whole set.}
\label{tab:mechfull}
\end{table}

\section{Evaluation-Budget Sweep}
\label{app:budget}
The budget sweep of Figure~\ref{fig:budget} re-tokenizes every evaluation response before applying each budget cutoff. The penalty rises to its peak at the $8$K budget and never closes as the budget extends to the $32$K cap, which rules out a budget-artifact reading. Across the three training seeds $\{42,123,7777\}$ the penalty at the $32$K cap is $15.6$, $16.6$, and $16.4$ points, the derivation deficit is $9.5$, $9.1$, and $7.3$, and the sweep peaks at the $8$K budget in every seed, so the shape is robust across seeds. The Qwen3-4B curve is uniformly higher, mirroring its larger overall penalty. The significance tests for these gaps are in Appendix~\ref{app:stats}.

\section{Chain-Population Split Robustness}
\label{app:twobytwo}
Table~\ref{tab:split} shows the carrier split of Section~\ref{sec:carrier} replicating at two further training seeds, the \textsc{Leaked-Late}-minus-\textsc{Leaked-Early} gap staying decisively positive at all three. The \textsc{Leaked-Early} model produces no extractable answer on $15.4\%$ of evaluation problems against $1.6\%$ for \textsc{Leaked-Late} and $1.2\%$ and $0.4\%$ for the blind arms, the termination pathology concentrating where the mechanism predicts.

The $438/497$ partition is not knife-edge. A strict answer matcher agrees with the default on $98.9\%$ of the $935$ leaked chains. Sweeping the answer-first threshold over $\{0.1, 0.2, 0.3\}$ shifts the early share only across $43.4\%$, $46.8\%$, and $49.6\%$. Eleven chains never state the answer inside the think block and are conservatively assigned to \textsc{Leaked-Late}.

The blind twins of $S_{\text{early}}$ problems are themselves answer-first $48.9\%$ of the time against $6.0\%$ for $S_{\text{late}}$, so the split partially reflects problem-level properties and motivates the blind control arms.

\begin{table}[t]
\centering
\small
\setlength{\tabcolsep}{5pt}
\renewcommand{\arraystretch}{1.15}
\begin{tabular}{lccc}
\toprule
Seed & \textsc{Leaked-Early} & \textsc{Leaked-Late} & Gap \\
\midrule
$42$   & 64.2 & 86.2 & $+22.0$ \\
$123$  & 71.8 & 85.0 & $+13.2$ \\
$7777$ & 66.0 & 84.0 & $+18.0$ \\
\bottomrule
\end{tabular}
\caption{Seed-level carrier arms, MATH-500 accuracy of the \textsc{Leaked-Early} and \textsc{Leaked-Late} fine-tunes at each training seed, with Gap the \textsc{Leaked-Late} minus \textsc{Leaked-Early} difference.}
\label{tab:split}
\end{table}

The decision rules are as follows.
\begin{enumerate}
\item The carrier hypothesis holds when the difference-in-differences interval excludes $0$ and the two blind arms differ by at most $3$ points.
\item The full-carrier outcome additionally requires \textsc{Leaked-Late} within $3$ points of the blind arms.
\item The partial-carrier outcome requires \textsc{Leaked-Late} at least $5$ points above \textsc{Leaked-Early} but more than $3$ points below blind.
\end{enumerate}

\section{Generation, Training, and Harness Details}
\label{app:details}
\paragraph{Generation.}
For the one-bit experiment (Section~\ref{sec:g1}) we sample $n{=}2$ chains for each problem at temperature $0.6$, top-$p$ $0.95$, top-$k$ $20$, with a $16{,}384$-token thinking budget, from Qwen3-8B. The \hint{} prompt appends the gold answer and requests the reasoning that reaches it, and the \nohint{} prompt is the standard solve instruction. We keep the first correct chain for each problem, with the final boxed answer verified by symbolic matching, intersect the problem sets solved under both conditions to obtain the $935$ matched problems, and fine-tune each condition on (question $\rightarrow$ chain) pairs with identical training inputs. These problems are drawn from the math split of our multi-domain SFT training corpus and are disjoint from the held-out MATH-500 test set on which all models are evaluated, so the comparison reflects generalization to unseen problems rather than memorization of evaluated items.

\paragraph{Training and evaluation.}
All SFT runs use DeepSpeed ZeRO-3 \citep{rajbhandari2020zero}, $1$ epoch, sequence length $10{,}240$, learning rate $2\times10^{-5}$, and effective batch size $48$, and all Qwen3-8B-arm models are full-parameter SFT of Qwen3-8B. Evaluation uses vLLM thinking mode at temperature $0.6$, top-$p$ $0.95$, top-$k$ $20$, and MATH-500 uses a $32{,}768$-token generation budget and symbolic answer matching. The base-model MATH-500 accuracies under this harness are Qwen3-8B $95.2$, Qwen3-4B $96.0$, Qwen3-1.7B $92.6$, DeepSeek-R1-Distill-Qwen-7B $79.6$, DeepSeek-R1-Distill-Llama-8B $77.8$, and DeepSeek-R1-Distill-Qwen-14B $88.2$ \citep{qwen3techreport,deepseekai2025deepseekr1}.

\paragraph{Gold-data baseline.}
A raw gold-data SFT baseline trained on the full multi-domain corpus scores $90.8$ on MATH-500 at the matched $10{,}240$-token budget, the gold-data reference that answer-blind self-distillation exceeds at the same budget.

\paragraph{Seed-level scores.}
Table~\ref{tab:seedlevel} reports the seed-level scores behind the three-seed means of Table~\ref{tab:main}(C). The MATH-500 blind-minus-leaked gaps of $15.6$, $16.6$, and $16.4$ average $16.2$, and the GSM8K and Minerva penalties average $4.9$ and $10.2$.

\begin{table}[t]
\centering
\small
\setlength{\tabcolsep}{5pt}
\renewcommand{\arraystretch}{1.1}
\resizebox{\columnwidth}{!}{%
\begin{tabular}{@{}llcccc@{}}
\toprule
Benchmark & Arm & Seed $42$ & Seed $123$ & Seed $7777$ & Mean \\
\midrule
MATH-500 & Blind  & 95.2 & 94.6 & 95.4 & 95.1 \\
         & Leaked & 79.6 & 78.0 & 79.0 & 78.9 \\
GSM8K    & Blind  & 95.6 & 95.2 & 95.4 & 95.4 \\
         & Leaked & 89.6 & 90.4 & 91.6 & 90.5 \\
Minerva  & Blind  & 48.2 & 48.5 & 49.3 & 48.7 \\
         & Leaked & 36.8 & 40.8 & 37.9 & 38.5 \\
\bottomrule
\end{tabular}
}%
\caption{Seed-level accuracies of the headline blind and leaked Qwen3-8B fine-tunes on the three math benchmarks whose three-seed means appear in Table~\ref{tab:main}(C).}
\label{tab:seedlevel}
\end{table}

\paragraph{Length-matched control.}
For the length-matched control, leaked training chains run about $5{,}800$ characters shorter than their blind twin, with $732$ of $935$ shorter. The length-matched blind arm scores $95.0$, within $0.2$ to $0.4$ points of the full blind condition and $15.4$ points above the leaked arm.

\section{Additional Details for the Generalization Experiments}
\label{app:beyond}
This section expands the setups and secondary statistics for the relaxations of Section~\ref{sec:beyond}, and all headline numbers appear in the main text.

\paragraph{Third-family point.}
For the independent third family of Section~\ref{sec:predict} we use NVIDIA's Llama-3.1-Nemotron-Nano-8B, a from-scratch Llama lineage distinct from Qwen and DeepSeek. Generation needs the system instruction \emph{detailed thinking on} to elicit a think block, after which the standard solve harness applies unchanged, and the fine-tuned arms emit a think block and a boxed answer on their own at evaluation. On $855$ matched problems the answer-first rates are $22.1\%$ blind and $40.0\%$ leaked, a signature of $\Delta\afr=+17.9$. Full SFT of each arm under the same recipe and harness as the other points gives blind $92.2$ against leaked $74.6$, a penalty of $+17.6$, within $0.2$ points of the $+17.4$ predicted from the six-model fit before training. The other held-out predictions were fixed from a four-model fit for Qwen3-1.7B and DS-R1-Distill-Qwen-14B and from a seven-model fit for the fourth-family GLM-Z1-9B. Table~\ref{tab:predict} collects the predicted and observed pairs, every one inside its band of about $\pm 9$ points.

\paragraph{Cross-student transfer.}
We fine-tune Qwen3-1.7B across three seeds on the Qwen3-8B-generated \nohint{} and \hint{} corpora built from the same $935$ problems, filter, and recipe. Table~\ref{tab:transfer} reports the resulting $+6.9$-point mean penalty, whose seed penalties are $16.8$, $-0.2$, and $4.0$, with the dispersion reported in Appendix~\ref{app:stats}. The small-student transfer therefore yields a positive penalty on average but is the least stable setting we test, in contrast to the seed-robust cross-family transfers. We read the average penalty as tracking the \emph{teacher's} $\Delta\afr$ of $+20.8$ rather than the student's $+26.2$, consistent with leakage being a property of the data.

\begin{table}[t]
\centering
\small
\setlength{\tabcolsep}{5pt}
\renewcommand{\arraystretch}{1.1}
\resizebox{\columnwidth}{!}{%
\begin{tabular}{@{}lcccc@{}}
\toprule
Base & Answer-cond. & Blind-only\,$\uparrow$ & STaR-rescue\,$\uparrow$ & Cost\,$\downarrow$ \\
\midrule
Qwen3-8B   & 45.8\% & 95.2 & 91.2 & \textcolor{red}{$4.0$} \\
DS-Qwen-7B & 27.6\% & 92.2 & 90.7 & \textcolor{red}{$1.5$} \\
\bottomrule
\end{tabular}
}%
\caption{STaR-replica rescue cost on MATH-500. Each corpus adds answer-hint rescues of the blind failures to the blind correct chains, so a fraction of it is answer-conditioned. Blind-only trains on the blind chains alone, STaR-rescue on the combined corpus, both filtered for correctness. Cost is Blind-only minus STaR-rescue. The DS-Qwen-7B row is a two-seed mean.}
\label{tab:star}
\end{table}

\paragraph{STaR-replica regime.}
We combine blind correct chains with answer-hint rescues of the failures, forming a corpus that is partly answer-conditioned, and Table~\ref{tab:star} reports the cost. A linear dose account would predict a cost proportional to the answer-conditioned fraction, and the measured $4.0$-point cost is real but somewhat sub-linear in that dose, plausibly because the rescued problems also contribute genuinely new content, which is the STaR rationale for rescuing them.

\begin{table}[t]
\centering
\small
\setlength{\tabcolsep}{8pt}
\renewcommand{\arraystretch}{1.1}
\begin{tabular}{@{}lcc@{}}
\toprule
Reference-echo rate (\%) & Blind & Leaked \\
\midrule
Early \textit{\footnotesize (first $20\%$ of reasoning)} & \cellcolor{blue!6}1.3 & 13.0 \\
Ever \textit{\footnotesize (anywhere in the chain)}      & \cellcolor{blue!6}12.2 & 92.1 \\
\bottomrule
\end{tabular}
\caption{Reference-echo rates of the blind and leaked code chains on the $379$ matched MBPP training pairs. A chain echoes when its reasoning contains a verbatim fragment of the reference solution, counted within the first $20\%$ of the reasoning (early) or anywhere in the chain (ever). The tinted column is the answer-blind arm.}
\label{tab:codeecho}
\end{table}

\begin{table}[t]
\centering
\small
\setlength{\tabcolsep}{5pt}
\renewcommand{\arraystretch}{1.1}
\resizebox{\columnwidth}{!}{%
\begin{tabular}{@{}lccc@{}}
\toprule
Setting & Blind\,$\uparrow$ & Length-matched\,$\uparrow$ & Leaked\,$\uparrow$ \\
\midrule
GLM teacher \textit{\footnotesize (MATH-500)} & \cellcolor{blue!6}93.0 & 90.9 & 82.8 \\
Code \textit{\footnotesize (MBPP+ pass rate)} & \cellcolor{blue!6}70.8 & 70.7 & 59.2 \\
\bottomrule
\end{tabular}
}%
\caption{Length-matched controls for the cross-family and code transfers, MATH-500 accuracy for the GLM row and MBPP+ pass rate for the code row. Each control truncates every blind chain to its matched leaked chain's length and retrains the student under the same recipe. Every code cell and the length-matched cells are means over the two seeds $\{42,7777\}$ of the control, and the GLM Blind and Leaked cells reproduce the corresponding arms of Table~\ref{tab:transfer}.}
\label{tab:lenmatch}
\end{table}

\paragraph{Code domain.}
From the $474$ MBPP training problems, with the test split unused, Qwen3-8B generates solutions blind and with the reference shown. Both arms keep only chains whose extracted code passes all asserts, and we train on the $379$ matched pairs. Table~\ref{tab:codeecho} reports the chain-level signature, whose early-echo gap of $\Delta_{\text{echo}}=+11.7$ points is the code-domain analogue of the early-answer behavior in math. The MBPP+ penalty, over its $378$ problems, and the HumanEval+ transfer penalty, over its $164$ problems, are three-seed means reported in Table~\ref{tab:transfer} and positive in every seed, with the MBPP+ seed values in Appendix~\ref{app:stats}. Table~\ref{tab:lenmatch} reports a length-matched control that truncates each blind chain to its matched leaked length, so the control chains run if anything shorter, at a mean $5{,}954$ characters against the leaked $8{,}378$. Its seed scores are $70.9$ and $70.4$ over seeds $\{42,7777\}$, far above the matched leaked arm, which shows content rather than length drives the code penalty.

\paragraph{Multiple-choice benchmarks.}
The same headline blind and leaked Qwen3-8B fine-tunes evaluated on multiple-choice benchmarks show no penalty across three seeds, consistent with the harm riding on multi-step derivation. On GPQA-Diamond, over its $198$ problems, the three-seed blind and leaked means are $37.6$ and $40.2$ against a base of $41.9$, and on a $2{,}000$-problem MMLU subset they are $54.8$ and $54.0$ against a base of $54.9$. Both arms sit near base and neither difference is distinguishable from zero, so on tasks that need little derivation the answer-leakage penalty is absent, and the seed-level values appear in Appendix~\ref{app:stats}.

\paragraph{Cross-family length control.}
For the GLM-Z1-9B teacher we repeat the length-matched control of Section~\ref{sec:g1}, truncating each blind chain to its matched leaked length and retraining a Qwen3-8B student under the same recipe. Table~\ref{tab:lenmatch} reports the arm, whose seed scores of $90.8$ and $91.0$ over seeds $\{42,7777\}$ sit near the untruncated blind arm and far above the leaked one, so the cross-family gap is content, not length.

\paragraph{Headroom regime.}
On DeepSeek-R1-Distill-Qwen-7B, whose base is $79.6$, the STaR-replica arm adds answer-hint rescues of the blind failures, and Table~\ref{tab:star} reports the resulting answer-conditioned fraction and its $1.5$-point mean cost over two seeds, positive in both. The one-bit experiment on the same base, blind against leaked self-distilled chains over $348$ matched problems, scores blind $91.4/89.2/88.4$ for a mean of $89.7$ against leaked $86.4/80.2/80.6$ for a mean of $82.4$ over seeds $\{42,123,7777\}$. The penalties of $5.0/9.0/7.8$, a mean of $7.3$, stay positive in every seed even though blind training gains about ten points over the base.

\paragraph{Derive-first mitigation.}
Regenerating the leaked chains with the gold answer in context but an explicit instruction to derive from scratch and reveal the answer only at the end produces the derive-first arm of Table~\ref{tab:main}(A). Its answer-first rate sits between the leaked and blind arms, the shift Section~\ref{sec:anatomy} reports.

\section{Statistical Significance, Confidence Intervals, and Tests}
\label{app:stats}
This section is the sole home for all confidence intervals, bootstrap and difference-in-differences intervals, standard deviations, standard errors, and significance tests reported anywhere in the paper. Each block below covers one experiment, and the point estimates and descriptive ranges it annotates appear in the matching body section.

\paragraph{Shared tests.}
Accuracy gaps and the difference-in-differences endpoint use a paired bootstrap over the evaluation problems, and discordant-item asymmetries use McNemar's test.

\paragraph{One-bit experiment.}
For the one-bit experiment the discordant items split $86$ to $8$ in favor of the blind arm ($p<10^{-17}$), the paired-bootstrap $95\%$ interval on the penalty is $[12.2,19.2]$ points, and the three-seed standard deviations are $0.4$ for the blind arm and $0.8$ for the leaked arm. On the pooled MATH-500 Level~4--5 band the blind and leaked seed dispersions are $\pm 0.2$ and $\pm 0.8$ with non-overlapping seed ranges.

\paragraph{Answer position.}
The answer-position shift is significant by a two-sample Kolmogorov--Smirnov test ($D=0.291$, $p<10^{-34}$).

\paragraph{Budget sweep.}
In the budget sweep the both-completed derivation gap and the Qwen3-4B completed-only gap are significant by McNemar's test ($51$ to $7$, $p=2.4\times10^{-9}$, and $p=1.9\times10^{-26}$).

\paragraph{Carrier experiment.}
In the carrier experiment, the two leaked arms differ decisively ($129$ to $19$, $p=2.8\times10^{-21}$) while the two blind controls do not ($p=0.48$), the difference-in-differences carries bootstrap $p<10^{-4}$ with a single-seed $95\%$ interval of $[16.4,25.8]$, and against their own blind twins the answer-first and derivation-first arms lose $29.8$ ($p=1.6\times10^{-38}$) and $8.6$ ($p=6.0\times10^{-8}$) points. For Qwen3-4B the leaked arm splits the discordant items $149$ to $3$ ($p=2\times10^{-40}$).

\begin{table}[t]
\centering
\small
\setlength{\tabcolsep}{4pt}
\renewcommand{\arraystretch}{1.15}
\resizebox{\columnwidth}{!}{%
\begin{tabular}{@{}lll@{}}
\toprule
Quantity & Value & Note \\
\midrule
Signature--penalty correlation $r$ & $0.960$ & exact permutation $p=0.0003$ \\
Leave-one-out range of $r$ & $[0.949,\,0.981]$ & dropping one model at a time \\
Fisher-z $95\%$ interval on $r$ & $[0.79,\,0.99]$ & wide at eight models \\
Partial corr., controlling base acc. & $0.91$ & permutation $p=0.0019$ \\
\quad bootstrap $95\%$ interval & $[0.79,\,1.00]$ & five effective degrees of freedom \\
Partial corr., controlling blind-arm acc. & $0.96$ & \\
Leave-one-family-out error & $2.9$ points & single-model families within $1.7$ \\
Excluding DS-R1-Distill-Llama-8B & $r=0.955$ & base-acc.\ partial rises to $0.93$ \\
\bottomrule
\end{tabular}
}%
\caption{Inferential statistics for the cross-model signature fit, the relation between the pre-fine-tuning $\Delta\afr$ signature of Eq.~\eqref{eq:afr} and the leakage penalty across the eight thinking models of Table~\ref{tab:signature}. The leave-one-family-out row fits on three families and predicts the entire held-out family, reporting mean absolute error in penalty points.}
\label{tab:sigstats}
\end{table}

\paragraph{Cross-model signature.}
Table~\ref{tab:sigstats} collects the inferential statistics for the cross-model signature fit of Section~\ref{sec:predict}. With only eight models the Fisher-z interval on $r$ is wide, and the held-out points share our generation and evaluation harness, so the fit should be read as a strong but small-sample trend rather than a precisely estimated law. The leave-one-family-out row shows the relation is not merely a Qwen-versus-DeepSeek family label, and the exclusion row shows the fit does not hang on its lowest-confidence anchor.

\paragraph{Code domain.}
The MBPP+ code-domain penalty is a three-seed mean, with blind $68.3$ (standard deviation $6.3$) against leaked $58.8$ (standard deviation $4.4$) and a mean penalty of $9.5$ points over seeds $\{42,123,7777\}$, with the three seed penalties $11.9$, $5.3$, and $11.3$. The HumanEval+ transfer is reported at $1.6$ standard errors.

\paragraph{Cross-student transfer.}
The cross-student transfer penalty has a seed standard deviation of $8.9$.

\paragraph{Multiple-choice benchmarks.}
On the two multiple-choice benchmarks the three-seed blind-minus-leaked means are $-2.6$ points on GPQA-Diamond and $+0.8$ on MMLU, so the leaked arm is marginally ahead on GPQA-Diamond and marginally behind on MMLU, and neither sign is distinguishable from zero across seeds $\{42,123,7777\}$, so we report no degradation rather than a reversed effect.

\paragraph{Off-distribution AIME.}
The off-distribution AIME penalty is a three-seed mean of $27.2$ points, with blind and leaked standard deviations of $5.8$ and $5.4$ over seeds $\{42,7777,123\}$.

\paragraph{Phi-4-reasoning.}
Phi-4-reasoning is the single thinking model tried and set aside for measurement reasons (Section~\ref{sec:predict}). It carries the largest signature in the set at $\Delta\afr=+45.5$, and the frozen six-model fit predicts a $47.4$-point penalty. Both trained arms, however, collapse to under $5\%$ MATH-500 accuracy, boxing an answer for only $3$ and $31$ of $500$ problems while rambling past the $32$K budget in degenerate loops with a mean response over $50$K characters. The realized gap is therefore degenerate rather than a valid penalty, and the point is discarded rather than graded favorably. We report the fit as an interpolation over the measured $\Delta\afr$ range rather than an extrapolation that this single unanchored extreme could confirm or distort.

\section{Reproducibility}
\label{app:repro}
\paragraph{Release contents.}
We release the code and data needed to regenerate every reported number. The release provides the exact generation prompts for all five conditions, the matched $935$ problem identifiers together with their answer-blind and answer-leaked chains, the generation metadata for each chain, the full training configuration, and the evaluation harness.

\paragraph{Correctness-critical components.}
The symbolic answer checker that decides which generated chains pass the keep-correct filter and which evaluation answers score as correct is the one correctness-critical component. Its dependencies are pinned to exact versions rather than version floors, so the keep-correct and evaluation decisions reproduce exactly. The release also exposes the answer-position matcher whose threshold-$0.2$ answer-first rate defines the chain-level signature.

\paragraph{Traceability.}
A manifest maps every benchmark, condition, and seed cell to the stored evaluation record that produced it, so each table entry is traceable to its source. The one setting a reader must pass explicitly is the training sequence length $10{,}240$, which every run uses. The held-out penalty predictions of Section~\ref{sec:predict} are the values we recorded before training each held-out model. The released code recomputes the cross-model fit from the stored evaluation files, so it reproduces the signature relationship and the leave-one-family-out error rather than re-emitting the frozen point predictions.

\paragraph{Neutral-arm comparability.}
The two neutral prompt-ablation arms in Table~\ref{tab:main} are trained with a numerically equivalent memory optimization and are reported as three-seed means. A matched answer-blind arm trained under that optimization scores $96.2$ at seed $42$, within about one point of the standard answer-blind reference, so the optimization does not materially shift accuracy and the two neutral arms are comparable to the standard answer-blind condition.

\bibliography{references}

\end{document}